\documentclass[letterpaper, 10 pt, conference]{ieeeconf}  

\usepackage{comment}
\usepackage{graphicx}
\usepackage{subfigure}
\usepackage{color}
\usepackage{xfrac}
\usepackage{array}
\usepackage{gensymb}
\usepackage{hyperref}
\usepackage{soul}
\usepackage{tabularx,ragged2e,booktabs,caption}
\usepackage[textsize=tiny]{todonotes}
\usepackage[detect-all,separate-uncertainty=true,multi-part-units=single]{siunitx}

\IEEEoverridecommandlockouts                              

\title{\LARGE \bf
Neuromorphic force-control in an industrial task: validating energy and latency benefits
}

\author{Camilo Amaya$^{a}$, Evan Eames$^{a}$, Gintautas Palinauskas$^{b}$, Alexander Perzylo$^{a}$, Yulia Sandamirskaya$^{c}$,\\Axel von Arnim$^{a}$
\thanks{$^{a}$~Department of Neuromorphic Computing, fortiss -- Research Institute of the Free State of Bavaria, Munich, Germany
        {\tt\small <LastName>@fortiss.org}
        }
\thanks{$^{b}$~The author was with the Department of Neuromorphic Computing, fortiss. He is now with Robominds GmbH, Munich, Germany
        {\tt\small gpa@robominds.de}
        }
\thanks{$^c$~ZHAW Zurich University of Applied Sciences/Intel Neuromorphic Computing Lab
        {\tt\small yulia.sandamirskaya@zhaw.ch}
        }
}

\begin{document}

\maketitle
\thispagestyle{empty}
\pagestyle{empty}

\begin{abstract}

As robots become smarter and more ubiquitous, optimizing the power consumption of intelligent compute becomes imperative towards ensuring the sustainability of technological advancements. Neuromorphic computing hardware makes use of biologically inspired neural architectures to achieve energy and latency improvements compared to conventional von Neumann computing architecture. Applying these benefits to robots has been demonstrated in several works in the field of neurorobotics, typically on relatively simple control tasks. Here, we introduce an example of neuromorphic computing applied to the real-world industrial task of object insertion. We trained a spiking neural network (SNN) to perform force-torque feedback control using a reinforcement learning approach in simulation. We then ported the SNN to the Intel neuromorphic research chip Loihi interfaced with a KUKA robotic arm. At inference time we show latency competitive with current CPU/GPU architectures, and one order of magnitude less energy usage in comparison to state-of-the-art low-energy edge-hardware. We offer this example as a proof of concept implementation of a neuromoprhic controller in real-world robotic setting, highlighting the benefits of neuromorphic hardware for the development of intelligent controllers for robots.

\end{abstract}
\IEEEpeerreviewmaketitle

\section{INTRODUCTION}

Moving towards the integration of intelligent robots in daily life, sustainability considerations require stricter optimization of energy consumption not only for actuation, but also for the computing involved in robot control. The current energy requirements of GPUs and CPUs either limit the scale and ``intelligence" of edge computation, or require high-latency communication protocols for cloud computing. While moving an AI system such as IBM Watson or ChatGPT to the edge requires killowatts of power for continuous inference\footnote{Ignoring training energy costs.} \cite{Greenemeier13,Brown20}, the human brain accomplishes much more complex behaviour at a tiny fraction of this \cite{Balasubramanian21}.  Biological inspiration therefore continues to influence both algorithm and hardware development in the field of neuromorphic computing, with an increasing number of applications in robotics. 

Neuromorphic hardware refers to a novel hardware architecture that uses principles of computing in biological brains and neural systems and has been shown to drastically improve latency and energy usage for many computing tasks \cite{Christensen22}, in particular ones that rely on recurrent, temporal, and sparse computation, as often the case in motion planning and control \cite{Davies21}. Typically, following biological inspiration, neuromorphic processors  realise in hardware so called Spiking Neural Networks (SNNs) \cite{Hodgkin52,Arthur11}. These differ from more ubiquitous Artificial Neural Networks (ANNs) in better exploitation of sparsity and recurrency through asynchronous multi-core processing (there is no global clock) and statefull neurons. Moreover, abundance of local memory enables efficient continual learning, along with the neural state updates. In recent years there has been an increasing interest in using neuromorphic hardware and SNNs for their advantages in energy and time efficiency in robotics --- a domain referred to as ``neurorobotics". We have proposed a roadmap for future development and deeper inspiration from biology for such systems \cite{Sandamirskaya22}.

Although a young area of research, neurorobotics has begun to see some first results. Examples include automotive object avoidance \cite{Mitchell17}, SLAM \cite{Tang19}, underwater propulsion \cite{Angelidis21}, drone control \cite{Vitale21,Peredes-Valles23}, and grasping force control \cite{Bao22}. See \cite{Bartolozzi22} for an overview. On the simulation front, the Neurorobotics Platform has been developed in the European Human Brain Project specifically as a sandbox for such applications \cite{Falotico17}.

However, as of yet the bulk of research has been carried out either partially or entirely within simulation \cite{Angelidis21,Volinski21,Alvarez21,DeWolf23,Palinauskas23}; with neuromorphic sensors paired with classical non-spiking hardware \cite{Muthusamy21,Ayyad23,Lawson23}; or with simple motion profiles (e.g. 1-dimensional movement) that do not constitute full use cases \cite{Stagsted20,Zaidel21,LinaresBarranco20,Vitale21,Bao22,Peredes-Valles23}. Other efforts have used customized neuromorphic circuits in which direct energy/latency measurements are not possible \cite{Lele22} or hybrid approaches that are only partially guided by neuromorphic hardware \cite{Ehrlich22}. See \cite{Aitsam22} for a full review. The consequence is that expected neuromorphic benefits cannot be inferred for real-world systems with practical applications.

To address this, we present a real-world neurorobotic system in which a robotic arm equipped with a force-torque (FT) sensor accomplishes an insertion task while fully controlled by neuromorphic hardware. The classic robotic insertion task has been chosen as it serves as the foundation for a number of related sub-tasks such as screw insertion, cable attachment, part assembly, etc. Additionally, the ``peg-in-hole" task has historically served as a benchmark for robotic integration of new algorithms, sensors, and hardware \cite{Brussel79,Nuttin97,Zhang17,Yoshimi94,Inoue17,Beltran-Hernandez20, Sileo24}, etc (see \cite{Xu19} for an overview).

The arm is trained in simulation using spiking reinforcement learning (RL). The trained network is then ported to an Intel Loihi neuromorphic research chip, which is connected to our KUKA robotic arm. We use sim2real techniques to accomplish insertion with the real robot. An accompanying video can be viewed at \href{https://tinyurl.com/y2szkbyb}{https://tinyurl.com/y2szkbyb}. 

This RL in simulation + sim2real technique is similar to that recently applied to drone and quadrupedal maneuvering using conventional ANNs \cite{Kaufmann20,Loquercio21,Miki22,Kaufmann23}. Our approach is similar, but realised with SNNs and deployed on a neuromorphic chip. We argue that, as the hardware matures, this could lead to energy efficient trained controllers for a broad deployment of intelligent robots. Our experiments have shown one order of magnitude less energy usage in comparison to current SotA non-neuromorphic edge-optimized hardware at a similar latency.

We highlight the following novel contributions of this work: this is the first real-world non-trivial robotic use-case (peg-in-hole insertion with an industrial robotic arm) fully guided by neuromorphic hardware. In this setting, energy and latency measurements of neuromorphic hardware were herein performed for the first time. We have found energy spent on running the network on the neuromorphic chip to be in the microjoule range. Moreover, this is the first neurorobotic system to incorporate a force-torque sensor.

\section{Methodology}
\label{Methodology}

\begin{figure*}[ht]
\begin{center}
\centerline{\includegraphics[width=0.99\textwidth]{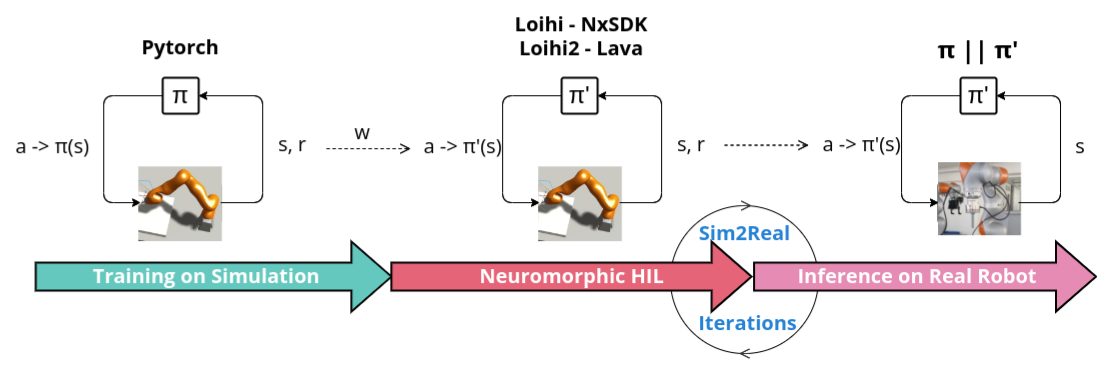}}
\caption{Learning and inference approach. HIL refers to `Hardware-In-Loop". An in-depth look at the third step can be found in Figure \ref{fig:FlowChart}.}
\label{fig:overview}
\end{center}
\end{figure*}

We first develop a simulation of the robotics setup, train the neural network based controller with spiking-RL using Pytorch, port the network to neuromorphic hardware, and finally run the trained policy on the real robot, applying various techniques to address the Sim2Real gap. A visualization of these steps is shown in Figure~\ref{fig:overview}.

\subsection{Simulation Setup}
\label{Simulation Setup}
\begin{figure}[ht]
\vskip 0.2in
\begin{center}
\centerline{\includegraphics[width=0.7\columnwidth]{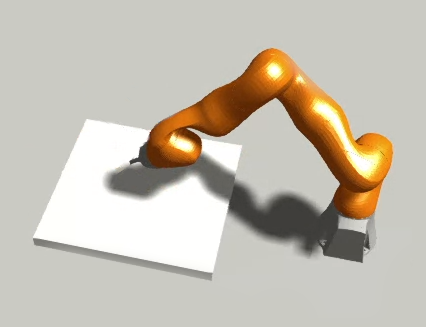}}
\caption{Neurorobotics simulation setup.}
\label{fig:nrp}
\end{center}
\vskip -0.2in
\end{figure}

The setup is simulated within the Neurorobotics Platform (NRP) \cite{Falotico17}, developed as part of the Human Brain Project \cite{Knoll16} and based on ROS \cite{Quigley09} and Gazebo \cite{Koenig04}. A robotic arm with 7 DOF mounted to a table is modelled. The end-effector consists of a cylindrical peg with a flat end and a 6-axis FT sensor. The table consists of a board with a circular hole. The peg can fit in the hole with a clearance of $\approx$~\SI{1}{\milli\meter}, and is initialized as per a Gaussian distribution centered on the hole with $\sigma =$~\SI{2}{\centi\meter}. This setup closely resembles the one used by previous teams (i.e. \cite{Inoue17}). The setup is shown in Figure~\ref{fig:nrp}. The full details of the simulation setup and underlying programming have been outlined in a separate complimentary publication \cite{Amaya23}. It primarily explores the implementation of RL within the NRP.

\subsection{Reinforcement Learning}
\label{Reinforcement Learning}

The simulated robotic arm is trained to find the hole using the spiking reinforcement learning technique described in \cite{Tang20}: a Population-encoded Spiking Actor Critic method (PopSAN). We base our state/action convention on that previously used by other teams in non-SNN works \cite{Xu19}. The state space consists of $[\vec{x},\vec{\theta},\vec{F},\vec{\tau}]$: position, angle, forces and torques (relative to the end-effector). These are encoded into spikes using population encoding, in which neurons represent discrete real numbers. A continuous input can be encoded through multiple neurons spiking in various combinations. To arrive at this, in our case the spiking actor network was tuned such that each of the 13 dimensions of the state space were encoded by 10 neurons (for a total of 130). The policy returns an action consisting of $[\vec{x},\vec{\theta}]$: a new target cartesian position and angle for the end-effector (returned as spikes, and then decoded into 6 real values for the low-level controller\footnote{\label{note1} Notice that even though the robot has 7 DOF, the policy does not control the nullspace as it is not deemed relevant to the task. The nullspace could be used in future works to optimize other metrics, such as energy efficient joint control, but this was out of the scope of this project.}). Each of the values in the action space is similarly represented by 10 neurons. 

The network connected to the population-encoded neurons was a fully connected SNN composed of \textit{Leaky Integrate \& Fire} (LIF) neurons arranged into layer sizes: $130\times256\times256\times60$, and was implemented using Intel's NxSDK software library \cite{Davies21}. The critic networks are fully connected ANNs with ReLU activation functions for the hidden layers and hyperbolic tangent activation functions for the output layer. Both Q-networks had sizes $19\times256\times256\times1$. For further details regarding the networks and parameters used refer to \cite{Tang20}.

After experimenting with a number of reward functions for RL, we settled on the following dense engineered reward:
\begin{equation}
\label{Reward Function}
R = w_1||f_d - f||_2 + w_2||\tau_d - \tau||_2 + w_3||z_d - z||_2
\end{equation}
Here $f_d, \tau_d, z_d$ are the desired force, torque, and depth. Similarly $f, \tau, z$ are the measured values for these three quantities. The weights $w$ allow us to adjust the importance we assign to each of these terms. Recall that the goal is to learn insertion based on the force-torque profile experienced by the end-effector. This is the motivation behind the first two terms. The third term serves to quickly teach the arm to remain on the table. Without this term the number of training episodes becomes impractically large.

For our purposes, we take $f_d$ and $\tau_d$ to be zero (this encourages a gentle insertion with minimal forces and torques). $z_d$ is set to be \SI{-0.07}{\meter} --- the bottom $z$ coordinate of the hole. The $w$ terms are heuristically adjusted based on the resulting behaviour. Explicitly, assigning high weight to the force and torque terms (effectively penalties, with $f_d = \tau_d = 0$) makes the arm hesitant to touch the table. Conversely, assigning these terms low weight can lead to erratic behaviour and dangerous forces that could damage the real set-up. For the $z$ term, high weight leads to less exploration in favour of simply pushing down, and low weight leads to indifference to the actual insertion. We set weights $\vec{w} = [0.05, 0.05, 0.9]$.

\subsection{Neuromorphic-Hardware-in-the-Loop}
\label{Neuromorphic-Hardware-in-the-Loop}

The policy is initially trained on an SNN simulated in PyTorch. Simultaneously, SNNs with the same architecture are defined in the corresponding frameworks for different neuromorphic hardware platforms (NxSDK for Loihi 1 and LAVA\footnote{https://github.com/lava-nc} for Loihi 2 \cite{Orchard21}). The trained weights can then be loaded onto the equivalent networks. The models may differ, as Loihi uses 9-bit precision values to describe the synaptic weights and neuron states, while the simulation allows for full precision values (64-bit floats). However, the differences were found to be negligible for this application. More specifically, we used two Intel Loihi 1 research chips in a USB-stick form factor system called Kapoho Bay \cite{Davies18} (seen in Figure~\ref{fig:real robot}), and a single Loihi 2 chip (Oheogulch) accessed remotely through the Intel Neuromorphic Research Cloud. 

\subsection{Real Robot Setup}
\label{Real Robot}

\begin{figure}[ht]
\vskip 0.2in
\begin{center}
\centerline{\includegraphics[width=0.6\columnwidth]{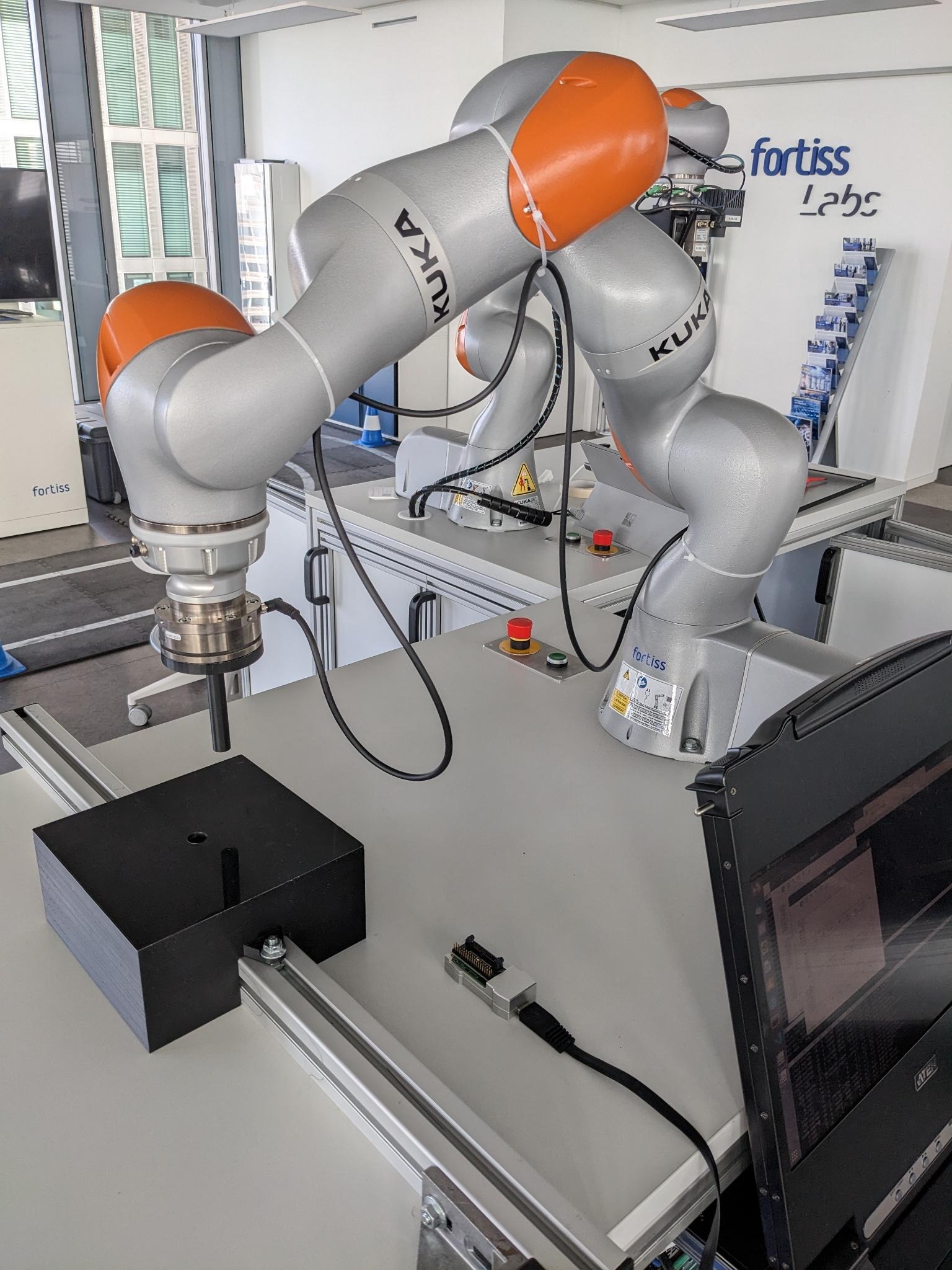}}
\caption{Demonstrator setup with a KUKA IIWA 7 R800 robot and attached cylindrical peg, black target box with hole, and Kapoho Bay containing two Loihi chips (on the tabletop).}
\label{fig:real robot}
\end{center}
\vskip -0.2in
\end{figure}

We use a KUKA IIWA arm with 7 DOF\footnote{See footnote \ref{note1}}. As in the simulation the end-effector consists of a solenoidal peg attached to a 6-axis FT sensor. The box containing the hole is 3D printed, as is the end-effector peg. An image of the experimental setup is shown in Figure~\ref{fig:real robot}.

\begin{figure}[ht]
\begin{center}
\centerline{\includegraphics[width=0.65\columnwidth]{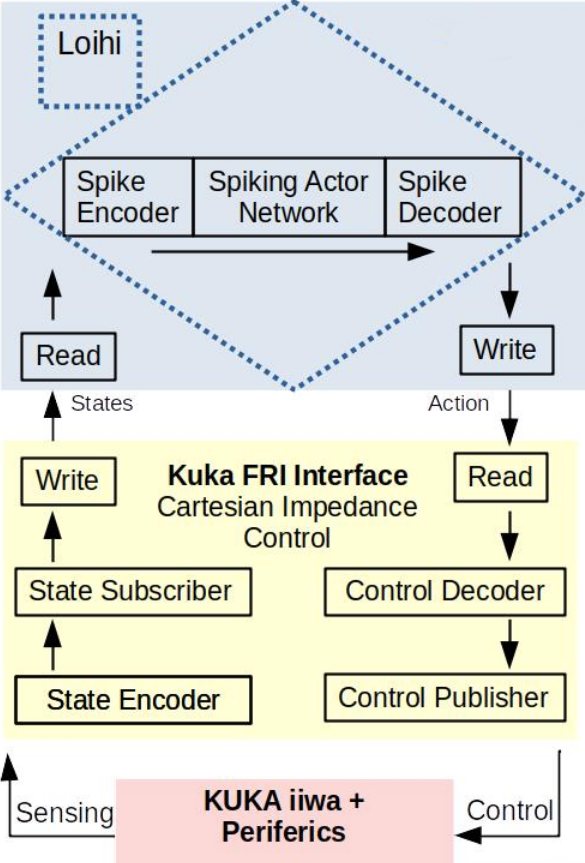}}
\caption{An overview of the communication mechanism between the high-level neuromorphic controller running on Loihi and the low level controller operating on the FRI.}
\label{fig:FlowChart}
\end{center}
\vskip -0.2in
\end{figure}

The spiking neural network trained in Section~\ref{Reinforcement Learning} controls the robot by providing a series of target poses. These targets can be connected through a smoothed path for reducing jerk in the motion, finally the path can be fed to a low-level impedance controller. The connection between these components is depicted in Figure~\ref{fig:FlowChart}.

\subsubsection{Controller}
\label{Controller}

A low level compliant controller was designed and implemented both for simulation as well as for the real robot. More precisely, a Cartesian Impedance Controller (CIC) was implemented and tuned using the methodology outlined in \cite{Albu-Schafer03}. The controller was implemented with the KUKA Fast Research Interface (FRI).

\subsubsection{Path Updating}
\label{Path Updating}

Small changes in the force-torque and positional values being measured from one moment to the next can trigger the policy to output a vastly different action from the previous one. Thus, the path must constantly be updated as new actions are received. For the compliant controller to accommodate smooth transitions between paths we use a fifth-degree polynomial for our path interpolation. This is chosen as it has been shown to reduce jerk, which, in turn, reduces manipulator wear and improves the trajectory accuracy and speed \cite{Macfarlane03}.

With the receipt of a new action from Loihi, the path is recalculated using the current pose, current velocity, current acceleration, and the final position (specified by the action). The final velocities and accelerations are taken to be zero.

\subsubsection{Sim2Real}
\label{Sim2Real}

To bridge the Sim2Real gap we employ both Domain Randomization and System Identification. These methods have been shown to work well for robotic manipulation tasks when first training in simulation \cite{Sievers22}. As the real coefficient of surface friction cannot be perfectly known (and can vary if switching end-effectors or surfaces) we allow the coefficient to take three values within the simulation $[0.34,0.38,0.42]$. From these, one is picked at the initialization of each training episode. These values are based upon repeated measurements of the real coefficient of friction on the printed surface. Multiple values assure robustness to variations in real surface friction \cite{Weng19}.

Additionally, the sensor force-torque noise distribution was characterized and employed during training. The noise profile follows a Gaussian distribution and is modelled on the slight changes in these values felt by the real FT sensor when not touching any surface.

Finally, we added a scaling factor to the received target orientations such that the amplitude of angular movements matches those in simulation. The mismatch was mainly due to unmodeled differences in the controller with respect to chattering between consecutive non-adjacent orientations. This could be acknowledge either by making the simulated controller more closely resemble the real equivalent, including safety measurements and interpolators to prevent actuator damage, or by adding a component in the reward function during training to prevent said chattering behaviour from being learned. However, for the sake of the task at hand we found that a simple scaling factor in the orientation provide a simpler solution.





\section{Results}
\label{Results}

\subsection{Training}

\begin{figure}[ht]
\vskip 0.2in
\begin{center}
\centerline{\includegraphics[width=0.95\columnwidth]{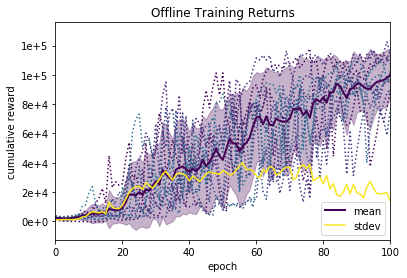}}
\caption{Learning curves showing the results of training with 20 random seeds, the mean performance and the standard deviation.}
\label{fig:learning_curve}
\end{center}
\vskip -0.2in
\end{figure}

The training (Figure~\ref{fig:overview}, left) was run for 100 epochs, with 500 episodes and 2000 interactions each. Figure~\ref{fig:learning_curve} shows the training profile of 20 different random seeds in terms of mean return value and corresponding standard deviation. We find a 100\% insertion success rate using a trained policy (across 50 runs, where the system is deemed to have failed if it does not insert within 30~s). Furthermore, there was no detected performance drop after porting the policy to real neuromorphic hardware while still using the simulated robotic arm (Figure~\ref{fig:overview}, center). Even though the models differ slightly due to quantization effects, these differences represent only negligible behavioural variations and still lead to the same success rates.


\subsection{Real-World Performance}

When the policy is initially run on the real robot (Figure~\ref{fig:overview}, right), we find an insertion success rate of 0\% (across 10 runs, where the system is deemed to have failed if it does not insert within 30~s). After the Sim2Real techniques (Section~\ref{Sim2Real}) we find a 100\% insertion success rate across 50 runs. Using success rate to compare the simulated and real implementations is therefore not possible, so we instead use the time to insertion (shown in Table \ref{tab:time-to-insertion}). All values are calculated across 50 runs.

A similar median and minimum insertion time is observed. Note that the mean real insertion time is roughly double the simulated time. This is due to a small number of outlier runs in which the peg has trouble navigating the complex real friction profile and temporarily gets stuck (which also explains the maximum insertion time discrepancy). Yet, perfect friction modelling is not our goal, and median insertion  time on the order of a few seconds is comparable with SotA implementations (see, for example, Figure~16 in \cite{Sileo24}). We now move to the energy and latency.\\

\begin{minipage}{\linewidth}
\centering
\captionof{table}{Time to Insertion}
\vspace{0.2cm}
\begin{tabular}%
{>{\centering\arraybackslash}p{2cm}||%
   >{\centering\arraybackslash}p{2cm}|%
   >{\centering\arraybackslash}p{2cm}%
  }
\label{tab:time-to-insertion}
& Simulation & Real\\
\hline
Mean & 3.7~s & {\centering 8.4~s}\\
Median & 3.4~s & 5.3~s\\
Minimum & 2.5~s & 2.2~s\\
Maximum & 7.4~s & 28.1~s\\
\end{tabular}
\end{minipage}

\subsection{Latency \& Energy}
\label{Latency & Energy}
\begin{figure}[ht]
\vskip 0.2in
\begin{center}
\centerline{\includegraphics[width=0.75\columnwidth]{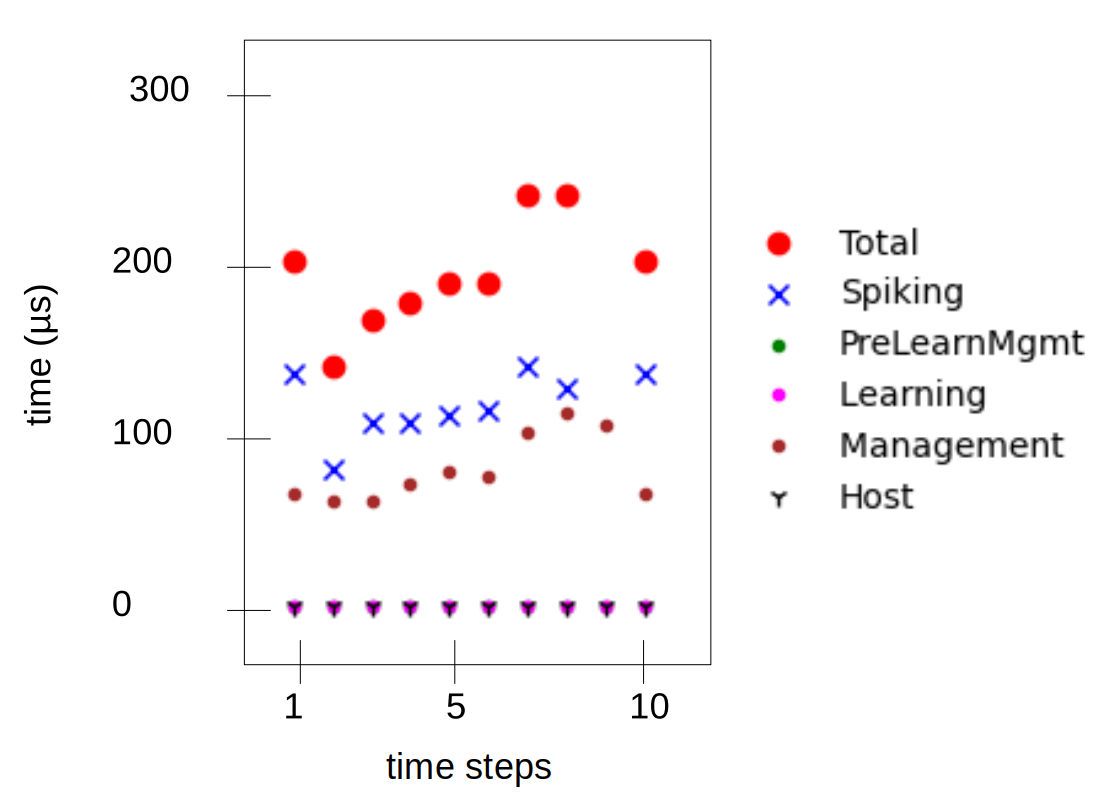}}
\caption{SNN execution time on Loihi 1 measured with state probes over 9 steps --- equivalent to one inference. The total is 1.8~\SI{}{\milli\second} (Table \ref{tab:performance_profiling}). Pre-learning management and learning are zero as the learning was off-line (and is additionally not relevant for inference).}
\label{fig:latency}
\end{center}
\vskip -0.2in
\end{figure}

Measuring the latency on the Loihi chip using the state probes we find the time profile shown in Figure~\ref{fig:latency}. We find slightly smaller latency on the Loihi 2 chip ($1.5 \pm 0.10$~\SI{}{\milli\second}). 

We find a per inference dynamic energy cost (removing the background energy cost of running the hardware) of $52\pm17$~\SI{}{\micro\joule}. This corresponds to a few tens of milliwatts. These results, in addition to comparisons with CPU and GPU values, are summarized in Table \ref{tab:performance_profiling}.\\

\begin{minipage}{\linewidth}
\centering
\captionof{table}{Energy and Time Profiling}
\vspace{0.2cm}
\begin{tabular}
{>{\centering\arraybackslash}p{2cm}||%
   >{\centering\arraybackslash}p{2cm}|%
   >{\centering\arraybackslash}p{2cm}%
  }
\label{tab:performance_profiling}
Hardware & $E_{\textrm{dynamic}}$ [\SI{}{\micro\joule}] & Latency [\SI{}{\milli\second}]\\
\hline
CPU $^\dagger$ & $3800$ & $1.4 \pm 0.1$\\
GPU~~~ & $\sim$ a few 100$^\ddagger$ & ~$2.0 \pm 0.1$*\\
Loihi 1** & ---  & $1.8$\\
Loihi 2~~~~ & $53 \pm 17$ & $1.5 \pm 0.1$\\
\end{tabular}

{\scriptsize$^\dagger$ CPU: 11th Gen Intel® Core™ i7-1165G7 @ 2.80GHz × 8.}\\
{\scriptsize$^\ddagger$ NVIDIA® / Pascal 256 CUDA cores @ 1300 MHz / Jetson TX2 (\cite{Lahmer22} Alg. 1 + personal communication with author)}\\
{\scriptsize * NVIDIA® / Mesa Intel® Xe Graphics (TGL GT2) / GeForce MX550}\\
{\scriptsize ** Note: Energy values are not available on Loihi 1 due to probe limitations.}\\
\end{minipage}

\section{Discussion}
\label{Discussion}

\subsection{Latency \& Energy}

Contrasting the recorded Loihi energy values with SotA non-neuromorphic edge computational hardware (Table~\ref{tab:performance_profiling}), the potential benefits are evident. Currently, optimized edge-hardware requires energy on the order of hundreds of \SI{}{\micro\joule} per inference for similar tasks/networks \cite{Lahmer22,Tang20b,Wisniewski22}, an order of magnitude more than neuromorphic hardware. Such hardware could therefore allow for more complex computations on autonomous robots, as well as orders-of-magnitude longer times between recharging. Additionally, non-neuromorphic edge-hardware often achieves low power consumption at the cost of latency (generally on account of cloud communication). For instance, \cite{Tang20b} cite latencies on the order of hundreds of milliseconds, compared to a few milliseconds by computing directly on Loihi (Figure~\ref{fig:latency}). It is also important to remember that Loihi is intended as a flexible research platform and has not been optimized for specific tasks.

Previous RL implementations of peg-in-hole (e.g. \cite{Inoue17,Luo18,Nuttin97}) generally invoke ANNs of a similar size to the SNN introduced here. Therefore, we should expect that the inference-time energy improvements over CPU and GPU (Table \ref{tab:performance_profiling}) apply here. Of course, inference (and computation in general) represents only one part of the entire robotic control system. For larger robots (10s of kg), the power requirements of the mechanical actuators are roughly one order of magnitude larger than the power requirements of computation \cite{Kashiri18}. Yet for sub-kilogram robots computation is likely to be a non-negligible source of power-consumption. As edge-robots become smaller and the tasks we expect them to perform become more complex, efficient compute will become even more imperative.

The measured energy and latency results are comparable to those reported in simulated neuromorphic use cases, hence validating the idea that the benefits of neurorobotic systems continue to hold when moved from simulations and simplistic use cases to more complex real-world applications. For example, \cite{DeWolf23} reports $\approx$~200~\SI{}{\micro\joule} and $\approx$~1~\SI{}{\milli\second} per inference using a slightly larger network on Loihi 1 for simulated robotic control. \cite{Vitale21} run a 4-layer SNN on Loihi 1D drone control. Although energy values are not reported, they do find a latency of 0.05~\SI{}{\milli\second} per step, corresponding to an equivalent of $\approx$~0.5~\SI{}{\milli\second} per inference. This value may be smaller than what we find on account of especially sparse data (coming from a neuromorphic camera). We noted that the latency values for CPU and GPU vary largely between devices. Indeed, \cite{DeWolf23} cite 15~\SI{}{\milli\second} latency on CPU, 10 times larger than our value. Ultimately we find similar latency between CPU and neuromorphic hardware.

It should be noted that Table \ref{tab:performance_profiling} refers to the computation times on-chip. On account of a known I/O bottleneck in Loihi~1 the robot-chip communication, which in theory should be negligible, ultimately increases the latency to $\approx$~16~ms. Although addressed in Loihi~2 \cite{Intel22}, we do not have an on-site physical chip, and therefore cannot measure the end-to-end latency on account of the cloud-communication cost. There is no reason, however, that the end-to-end latencies should be larger than what is quoted in Table \ref{tab:performance_profiling}. We also remind the reader that the energy per inference represents only one part of the overall energy required for computation. Loihi currently requires significant power for overhead operations (such as for spike I/O - handled by a separate FPGA). These are not included, as they are expected to be optimized with subsequent chip releases, to the point that the inference energy is primary (as was the case for classical chips).

\subsection{Application Specific / Peg-in-hole }

The insertion times (Table~\ref{tab:time-to-insertion}) are comparable between simulation and the real robot. We find quantization effects on Loihi (previously explored by \cite{Akl21}) to be negligible. We notice that, even when applying domain randomization in simulation training to vary the friction coefficient (Section~\ref{Sim2Real}) the movement behaviour still differs somewhat due to the complex friction profile between the peg and surface. Notably, on the real surface, the peg will often initially not move, and then jerk forward when the friction is overcome. This jerking was not captured in simulation, although insertion on the real robot is still successful regardless. Velocity and acceleration limits are the standard KUKA preset values. We do not expect varying the limits would affect the outcome, although this would need to be explicitly tested.

We had initially hypothesized that the arm, using the FT sensor and ability to move in angular space, would learn to feel the force-torque profile corresponding to being on the edge of the hole. We find that the trained policy is such that the angular movement is quite small ($\pm\approx 1$\degree{} for all three axes). This means that generally the peg is held approximately vertical. Additionally, during the exploration phase, the trained policy returns actions sending the end-effector back-and-forth into the positive and negative $x$ and $y$ directions (relative to the hole). Although the peg appears to sometimes react to arriving at the edge of the hole (by moving towards the hole centre), in a number of runs we certainly see the peg slipping into the hole, likely by accident, from this learned sweeping behaviour. However, in the case where the peg only slips part-way into the hole, the slight angular movements do appear to help it arrive at the bottom.

In a future work, it would be worthwhile exploring a larger policy network allowing for more complex behaviour, more robust noise modelling (to account for the above-mentioned jerky motion), and holes at different angles to make the sweeping motion insufficient for insertion.

\subsection{Neurorobotic Outlook}
\label{Neurorobotics}

We have demonstrated robotic insertion using a neuromorphic high-level controller as a proof of concept, meant to emphasize that neuromorphic computing is no longer simply up-and-coming, but rather sufficiently mature to move towards tackling real robotic use cases. This implementation is additionally intended to act as a software base for future neurorobotic development. 

Indeed, flexible research platforms such as Loihi can already be used as test-beds and prototyping tools for current non-research neuromorphic and mixed-signal devices with SoTA performance\footnote{eg. DYNAP-SE, ROLLS, Innaterra, etc.}. Neuromorphic hardware also allows for ``online'' and ``continuous'' learning unique to SNNs and inspired by the human brain which, although found not to be needed for successful insertion in our use case, we nonetheless hope to explore in a future publication.

Looking forward, it is worth mentioning that current actuators do not operate on spiking principles. With neuromorphic hardware and sensors already in existence, spike-based actuators would allow for neurorobots in which energy and latency is not lost to encoding and decoding spikes between neuromorphic and non-neuromorphic components. Additional research is here needed to unlock the true potential of neurorobots.

Ultimately, and as with the first RL peg-in-hole implementation, we hope that this first neuromorphic implementation contributes to an exciting new domain of robotics and automatization --- that of neurorobotics.

\section*{Code}
\begin{center}
Code available upon request.
\end{center}

\section*{Acknowledgements}

The research at fortiss was supported by the HBP Neurobotics Platform funded from the European Union's Horizon 2020 Framework Program for Research and Innovation under the Specific Grant Agreements No. 945539 (Human Brain Project SGA3).


\bibliographystyle{unsrt}
\bibliography{Amaya2024}

\end{document}